%% file: root.tex
\documentclass[10pt,twocolumn,letterpaper]{article}

\usepackage[pagenumbers]{cvpr}

\usepackage{graphicx}
\usepackage{amsmath}
\usepackage{amssymb}
\usepackage{booktabs}
\usepackage{comment}
\usepackage{bm}
\usepackage{xcolor}
\usepackage{bbm}
\usepackage{makecell}
\usepackage{enumitem}
\usepackage{stackengine}
\usepackage{nicematrix}
\usepackage{tikz}
\usepackage{multirow}

\usepackage[pagebackref,breaklinks,colorlinks]{hyperref}

\usepackage[capitalize]{cleveref}
\crefname{section}{Sec.}{Secs.}
\Crefname{section}{Section}{Sections}
\Crefname{table}{Table}{Tables}
\crefname{table}{Tab.}{Tabs.}

\def\cvprPaperID{***}
\def\confName{UNCV-W}
\def\confYear{2024}

\begin{document}

\title{Transient Fault Tolerant Semantic Segmentation for Autonomous Driving}
\author{
Leonardo Iurada$^{1}$ \hspace{0.5cm} Niccolò Cavagnero$^{1}$ \hspace{0.5cm} Fernando Fernandes Dos Santos$^{2}$ \\ Giuseppe Averta$^{1}$ \hspace{0.5cm} Paolo Rech$^{3}$ \hspace{0.5cm} Tatiana Tommasi$^{1}$ \vspace{2mm} \\
$^{1}$Politecnico di Torino, Italy \hspace{0.5cm} $^{2}$Univ Rennes, INRIA, France \hspace{0.5cm} $^{3}$Università di Trento, Italy\\
\texttt{\small \{leonardo.iurada, niccolo.cavagnero, giuseppe.averta, tatiana.tommasi\}@polito.it}\\
\texttt{\small fernando.fernandes-dos-santos@inria.fr} \hspace{0.5cm} \texttt{\small paolo.rech@unitn.it}}
\maketitle

\input{sections_abstract}

\input{sections_introduction}
\input{sections_related}
\input{sections_method}
\input{sections_experiments}
\input{sections_conclusions}

{\small
\bibliographystyle{ieee_fullname}
\bibliography{egbib}
}

\end{document}

%% file: sections_abstract.tex
\begin{abstract}
Deep learning models are crucial for autonomous vehicle perception, but their reliability is challenged by algorithmic limitations and hardware faults. We address the latter by examining fault-tolerance in semantic segmentation models. Using established hardware fault models, we evaluate existing hardening techniques both in terms of accuracy and uncertainty and introduce ReLUMax, a novel simple activation function designed to enhance resilience against transient faults. ReLUMax integrates seamlessly into existing architectures without time overhead. Our experiments demonstrate that ReLUMax effectively improves robustness, preserving performance and boosting prediction confidence, thus contributing to the development of reliable autonomous driving systems.
\emph{Code available at: \url{https://github.com/iurada/neutron-segmentation}}
\end{abstract}

%% file: sections_introduction.tex
\vspace{-5mm}
\section{Introduction}

Autonomous vehicles face significant challenges in perceiving and navigating complex environments. Reliable scene recognition models are crucial, especially for Advanced Driver Assistance Systems (ADAS) that must comply with functional safety standards like ISO 26262 \cite{ISO26262}. While deep learning has advanced capabilities such as obstacle detection and traffic sign recognition, certification of these components remains a challenge.
Recent research has focused on improving algorithmic robustness through domain generalization \cite{DGsegmentation,wei2023stronger}, anomaly detection \cite{shyamICCV23}, and open-set recognition \cite{Li_2023_ICCV}. However, hardware robustness is equally critical. Transient hardware faults, often caused by cosmic particles, can result in bit-flip errors \cite{Baumann2005,AhmadilivaniSurvey} that may lead to incorrect predictions and potentially fatal decisions in autonomous vehicles (see \cref{fig:teaser}).
Our work addresses this hardware vulnerability in the context of semantic segmentation, a key task in scene interpretation for autonomous driving. We aim to understand and mitigate the impact of hardware errors on this critical function.

\input{images_figures_teaser}

Ideal fault-resilient systems require low-latency and cost-effective strategies. However, current solutions involve expensive hardware or high-cost redundancy \cite{abaec12019NVDLA, dris3DMemoryHard2019}, exemplified by Tesla's Full Self-Driving Chip \cite{9007413}. Traditional error-correcting code (ECC) focuses on GPU memories rather than functional units \cite{Fratin2018, Sulivan2021NVIDIADDR}. Software-based strategies typically adapt classical techniques to neural networks, introducing significant time overhead \cite{fatUssama2020,sivaABFT2021}. Recent research on computer vision model reliability addresses limited datasets with simplified fault models \cite{Mahmoud2021,wasim2023textguidedresilience} or relies on reactive post-processing approaches \cite{amms_burel}, overlooking the model training process.

\textbf{With this work, we present to the computer vision community}: 
\begin{itemize}[leftmargin=*]
\vspace{-2mm}
\item the first fault tolerance analysis on deep learning-based semantic segmentation for autonomous driving. Our study leverages fault models obtained from physical experiments rather than standard synthetic ones. 
\vspace{-2mm}
\item a new hardening technique for deep convolutional segmentation models. We introduce the  activation function ReLUMax that allows monitoring the training phase and operates corrections at inference time with no latency.
\end{itemize}

Our experimental evaluation based on a fault injection campaign shows how the proposed solution reduces transient computational errors maintaining an accuracy close to that of the fault-free setting, significantly reducing the number of critical errors and presenting top model confidence.

%% file: images_figures_teaser.tex
\begin{figure}[t]
\centering
    \stackunder[5pt]{\includegraphics[width=.46\linewidth]{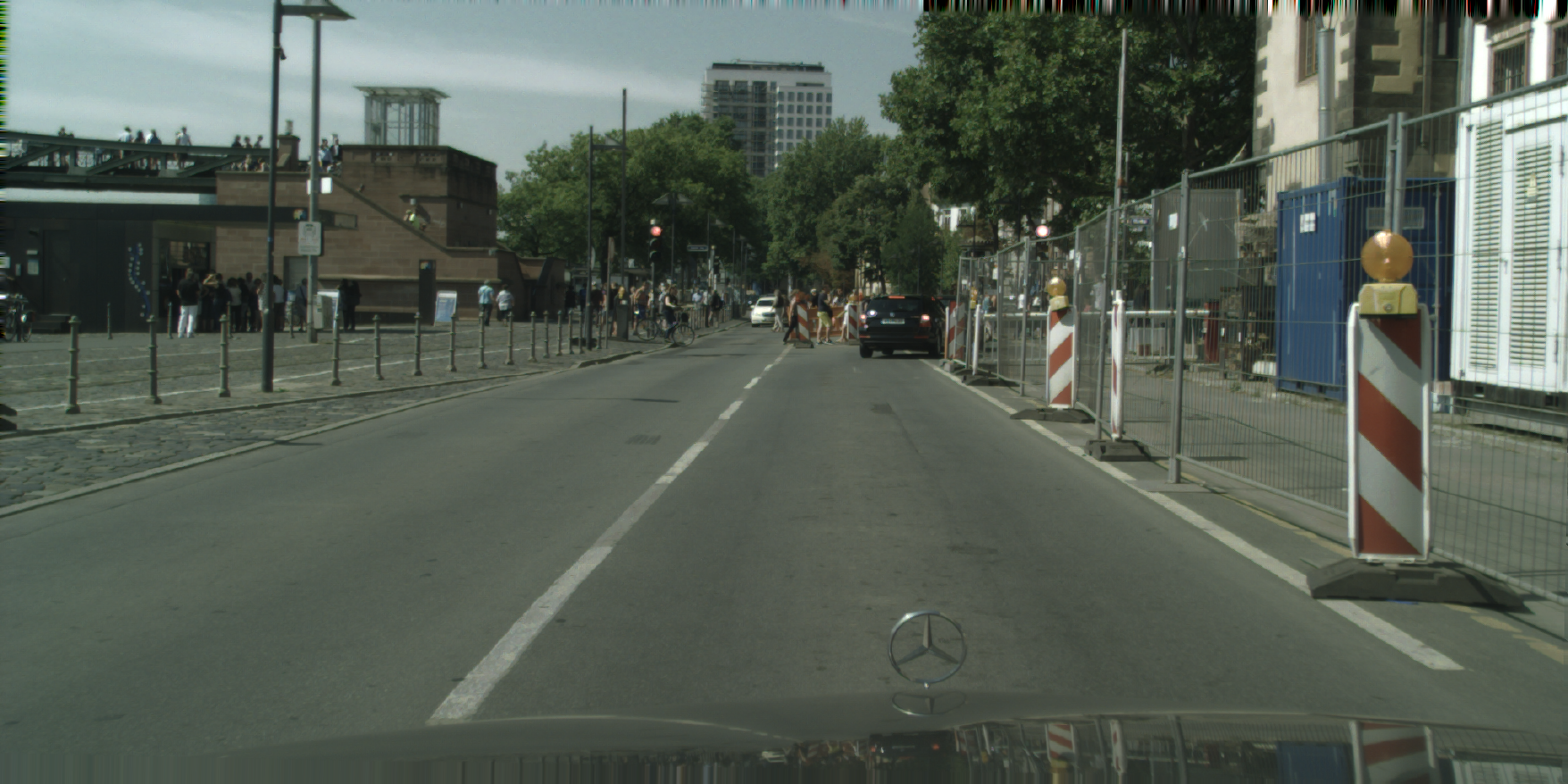}}{RGB Image}
    \hspace{12pt}
    \stackunder[5pt]{\includegraphics[width=.46\linewidth]{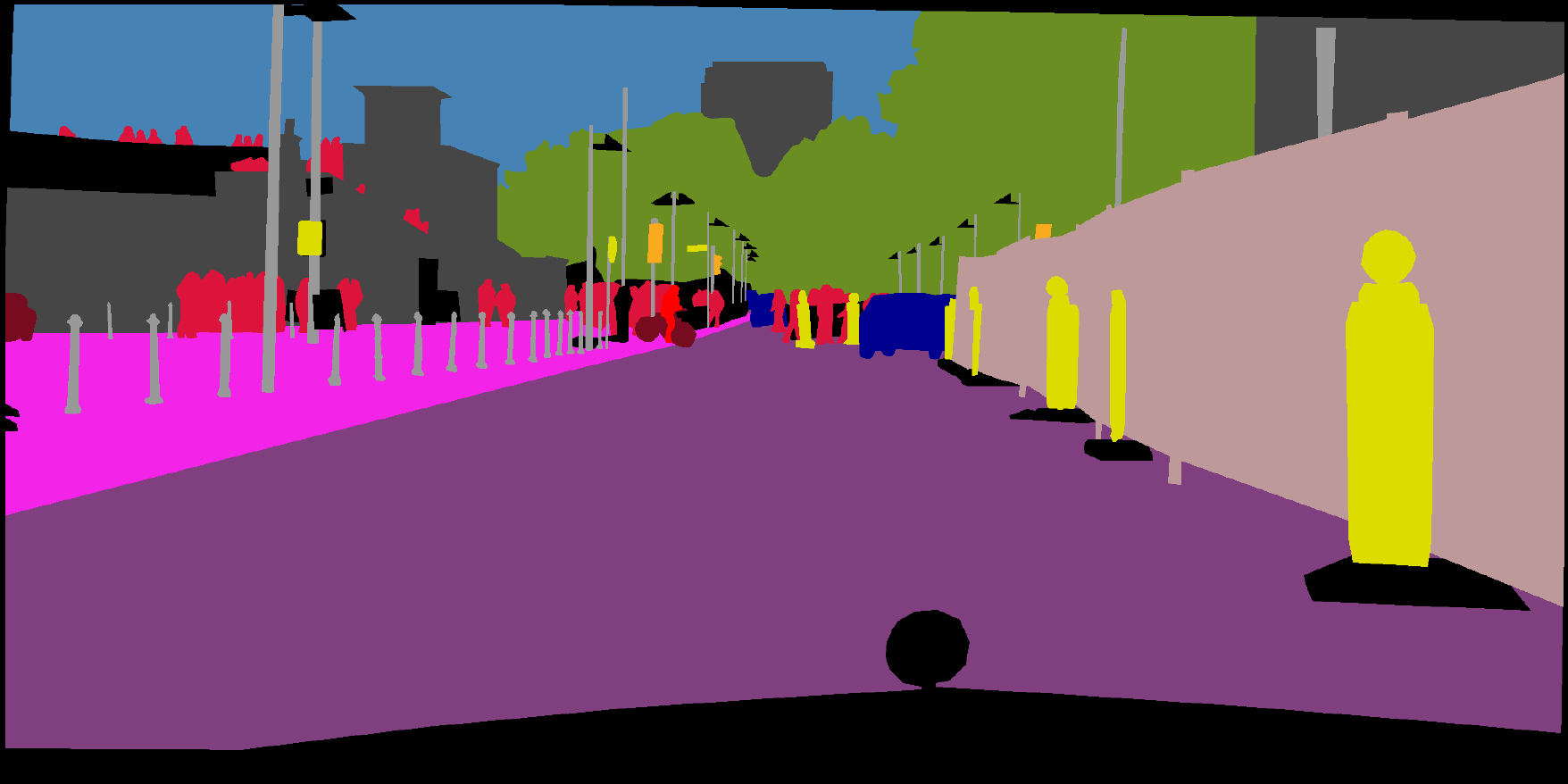}}{Ground Truth}
    \hspace{12pt}

    \vspace{12pt}
    \stackunder[5pt]{\includegraphics[width=.46\linewidth]{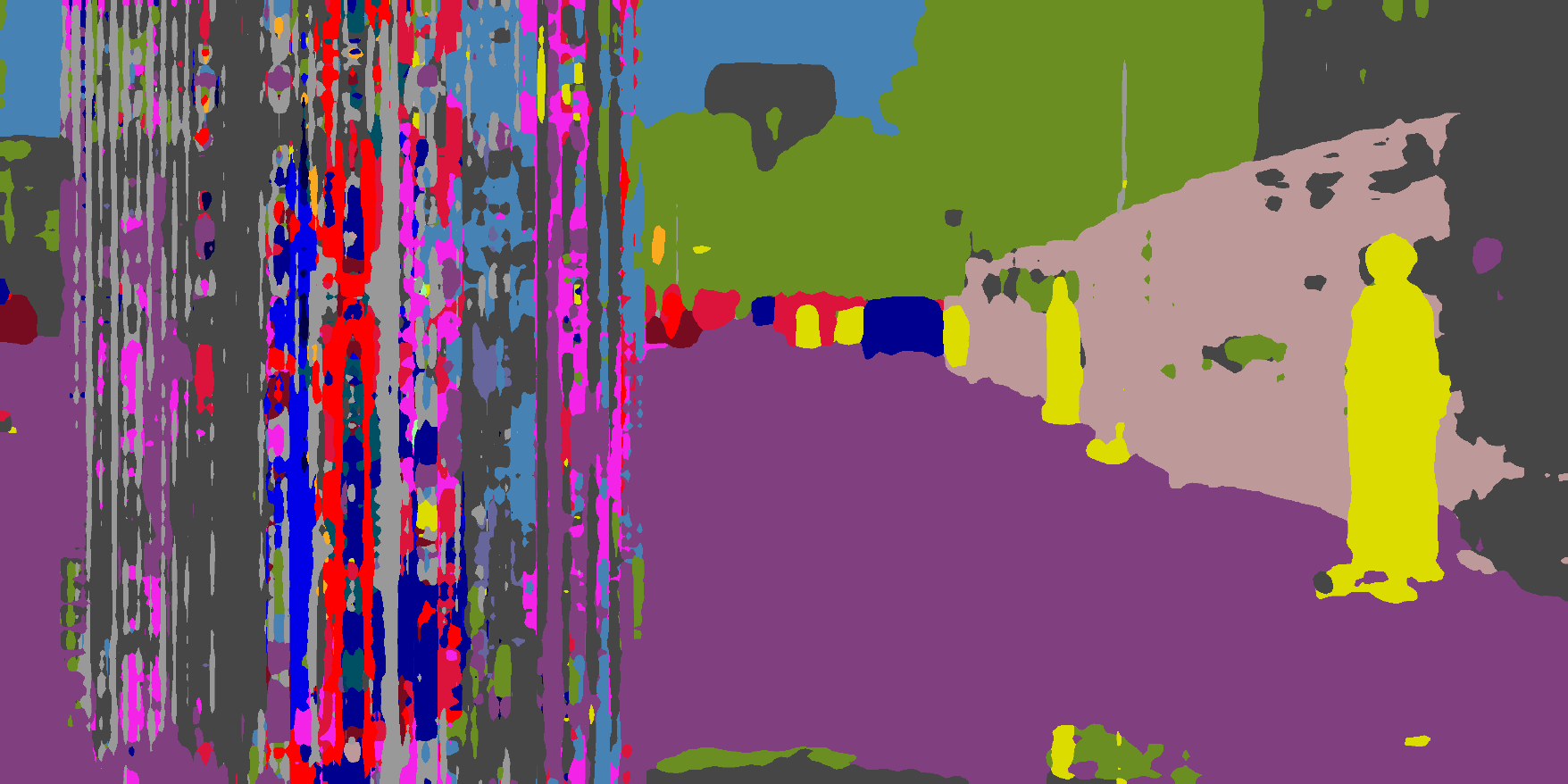}}{No Hardening}
    \hspace{12pt}
    \stackunder[5pt]{\includegraphics[width=.46\linewidth]{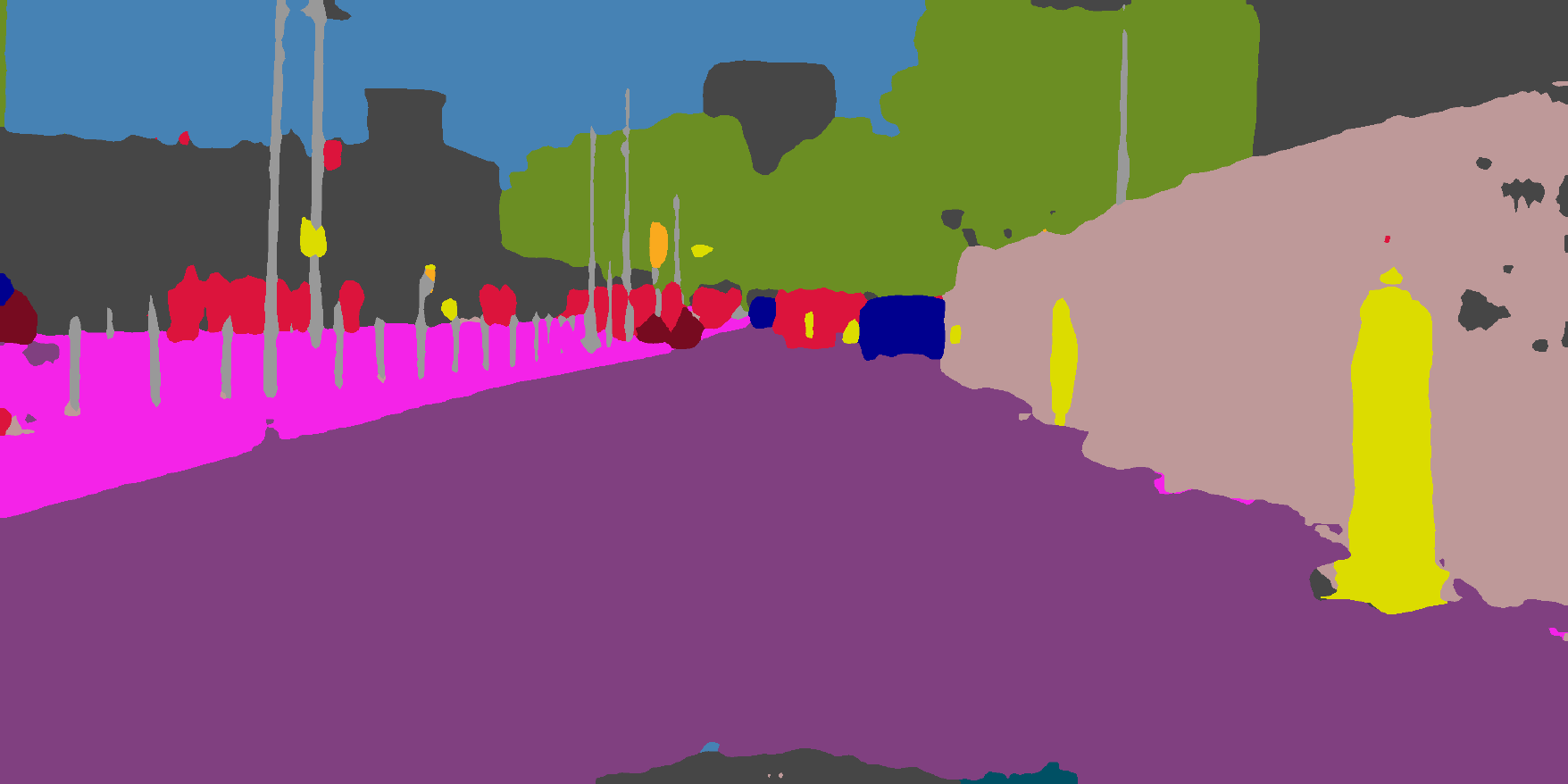}}{ReLUMax (Ours)}
    
    \caption{Semantic segmentation models may experience catastrophic output corruption under transient faults, rendering them unsafe for critical applications (No Hardening). To address this limitation, we propose a novel approach for automatically hardening activation functions at training time, without incurring in any additional cost (ReLUMax). Our method ensures robustness against transient faults, mitigating severe corruptions and significantly improving the system's trustworthiness.}
    \label{fig:teaser}
    \vspace{-5mm}
\end{figure}

%% file: sections_related.tex
\section{Related Works}

\noindent\textbf{Transient Faults and Hardening.}
Transient faults affect deep learning system reliability, with impact varying by network architecture and processing unit. Software hardening techniques to mitigate these effects are actively researched. Strategies include selective feature map duplication~\cite{Mahmoud2021}, convolution checksums~\cite{santosTR2019, Mittal2020, sivaABFT2021, libanoTopTC2023}, re-execution~\cite{rpdwc2022, Libano2019}, prediction ensembles~\cite{GaoEnsemble2020}, specialized pooling layers~\cite{santosTR2019, li2017SC}, and value attenuation~\cite{li2017SC,rangeRestrict}. Recent approaches designed for object classification combine fault-aware training, ReLU activation clipping, and batch normalization positioning~\cite{cavagneroIOLTS2022}. 
Only two previous works considered the task of semantic segmentation on fish-eye images. One proposed to improve fault tolerance  by calculating activation statistics in a post-processing stage to identify faulty values and apply zero masking on them \cite{amms_burel}.
The collected statistics capture a late snapshot of the model and do not value the dynamic nature of the training process. 
The second work \cite{burel2024relu6}, adopts a fixed ReLU activation clipping as~\cite{cavagneroIOLTS2022} but observed that it may hinder training convergence and reduce accuracy compared to unmodified models.

\noindent\textbf{Fault Models.}
Faults can affect any component of deep neural network hardware platforms, with increasing risks as transistors shrink. Fault models represent how faults manifest as incorrect states leading to prediction failures. Common synthetic abstractions include bit-flips in weights, activations, or convolution outputs \cite{amms_burel,burel2024relu6}. Recent studies on neutron beam exposure show faults can corrupt feature maps, affecting entire rows, columns, or blocks of tensors, with magnitudes reaching infinity or NaN~\cite{santosTR2019}. We adopt the strategy from~\cite{cavagneroIOLTS2022}, considering random combinations of feature map region variations.

%% file: sections_method.tex
\section{Method}
\noindent\textbf{Background.}
One tangible consequence of hardware transient faults is a notable alteration in the range of internal deep network values, often resulting in the emergence of excessively large activations. Existing strategies to tackle this problem include manually setting upper bounds based on the values of the neurons in each layer in the absence of faults \cite{li2017SC}.
In \cite{amms_burel} the authors proposed to collect the distribution of the Average, Minimum, Maximum, and Standard deviation (AMMS) of the activation values for each layer at the end of the training phase. They fix an error detection threshold for each statistic, identifying when it is one standard deviation beyond the minimum or maximal value. 
They show that if the average and minimum are both out of range, it is possible to reliably identify a fault, and mitigate it by masking values to zero.

We remark that deep neural networks are inherently wired to manage out-of-range activation values thanks to the ReLU functions, thus adding handcrafted procedures is clearly suboptimal. 
A clipped ReLU activation function was used in \cite{Hoang2020} to map high-intensity (possibly faulty) activation values to zero. The selected clipping threshold was refined with a dedicated fine-tuning algorithm but could be below the maximum activation value in the training phase, possibly modifying the error-free behavior of the network. 
In \cite{cavagneroIOLTS2022} the authors adopted ReLU6 from the literature on efficient deep learning models where the threshold of 6 was chosen to reduce the risk of overflow/underflow \cite{relu6} and demonstrated to produce the best accuracy-reliability trade-off in case of hardware-permanent faults \cite{Hoang2020}. Moreover, they exploited fault-aware training via a tailored data augmentation which mimicked the effects of hardware faults to learn patterns that are robust to fault-related noise.
Later, \cite{burel2024relu6} discussed how the use of ReLU6 for mitigating faults in segmentation is not without drawbacks.

\noindent\textbf{ReLUMax.} We propose to leverage a new version of the ReLU function to improve deep neural network fault resilience for semantic segmentation, while overcoming the limitations of existing approaches. In particular, we introduce ReLUMax, which builds upon the established ReLU-n concept \cite{relu6}, but with a key distinction: it dynamically computes the optimal clipping value for each feature map during the training process. Each ReLUMax activation function stores the observed maximal value (\ie a single floating-point number) from its own output during training and uses it at evaluation time as a trigger to clip activation values to zero.

%% file: sections_experiments.tex
\input{tables_simulated_extended}

\section{Experiments}

In this section we present our experimental analysis to assess the performance of ReLUMax as a hardening solution for fault-resilient semantic segmentation.

\subsection{Experimental Setting}
\noindent\textbf{Architecture.} We use DeepLabV3 \cite{deeplabv3} on a ResNet-50 backbone \cite{resnet}. By following standard practices, we start from a pre-trained model obtained on a subset of COCO \cite{lin2014coco}, using only the 20 categories that are present in the Pascal VOC dataset \cite{everingham2015pascalvoc}. 

\noindent\textbf{Datasets.} We run the semantic segmentation experiments on GTA5 \cite{richter2016gta5} and Cityscapes \cite{cordts2016cityscapes} datasets, following standard procedures described in \cite{chen2017deeplab, liu2021source} and \cite{deeplabv3, cordts2016cityscapes}, respectively. Both are large-scale datasets for urban scene understanding. The former consists of 24,966 synthetic images $1052 \times 1914$ with pixel-perfect annotations. The latter, contains 3,975 images $1024 \times 2048$ with fine-grained annotations of real-world road scenes, comprising up to 30 different categories.

\noindent\textbf{Transient Fault Injection.} 
We inject transient faults using the module from \cite{cavagneroIOLTS2022}. Errors appear as row or column-wise stripes or localized blocks within the feature maps. Corruption involves multiplying the output tensor with a uniformly sampled random value determining the error magnitude. This fault injection is stochastically applied to random layers during each forward pass.

\subsection{Semantic Segmentation Results}

We evaluate ReLUMax against five baselines: \emph{No Hardening}, \emph{Fault-Aware Training} \cite{cavagneroIOLTS2022}, \emph{ReLU6} \cite{cavagneroIOLTS2022, burel2024relu6}, \emph{ReLU6 + Fault-Aware Training} \cite{cavagneroIOLTS2022}, and \emph{AMMS} \cite{amms_burel}. We use mean Intersection over Union (mIoU) for evaluation and classify silent data corrupts (SDCs) according to \cite{amms_burel, burel2024relu6} as \emph{Masked}, if no bit-level difference is present in the output logits. \emph{No Impact}, if the predicted pixel-level categories are the same. \emph{Tolerable}, if less than $1\%$ of pixels in the output prediction are affected by the SDCs and no semantic class appears or disappears. They are \emph{Critical} SDCs otherwise.
Results are presented in \cref{tab:sim}. In fault-free scenarios, hardening methods don't significantly impact performance, though ReLU6 shows a slight decrease. With fault injections, ReLUMax proves most effective, followed by AMMS. Both use similar masking logic, but ReLUMax estimates clipping thresholds during training, generating superior fault estimates. 

\input{images_figures_maps_zoom}
\subsection{Qualitative Effect of Fault Injections}

As shown in \cref{fig:teaser}, the absence of hardening measures leads to substantial corruption from transient faults, posing serious safety risks for autonomous driving. Faults typically generate patterns where entire columns in layer outputs are perturbed. ReLUMax demonstrates a notable ability to mitigate fault influence, providing consistent predictions.
\cref{fig:maps_zoom} illustrates the worst-case scenario recorded at inference time on Cityscapes. Without hardening, significant performance degradation is observed. Fault-aware training without clipping leads to completely corrupted predictions. ReLU6 activation avoids complete corruption but overestimates the \emph{person} class (in red) in shattered blobs. Combining ReLU6 with fault-aware training or using AMMS doesn't resolve the issue, with the \emph{person} class missing entirely. ReLUMax provides proper localization of persons even in the worst case.

\input{tables_uncertainty}

\subsection{Uncertainty Analysis of Hardened Models}

While mIoU assesses pixel-level segmentation accuracy, it overlooks model confidence, which is crucial in real-world applications where uncertain predictions can have significant consequences. To address this, we analyze model confidence using predictive uncertainty \cite{ddu_uncertainty}, computed via softmax entropy from model outputs. We then evaluate this uncertainty using four metrics proposed in \cite{ddu_uncertainty, pavpu, PRR_deJorge}.
For the first three metrics we need to decompose the image in patches of size $w \times w$, with $w > 1$ and evaluate on them pixel accuracy and prediction uncertainty.
The results are then collected in a confusion matrix containing the number of patches that are accurate and certain $n_{ac}$, accurate and uncertain $n_{au}$, inaccurate and certain $n_{ic}$ and inaccurate and uncertain $n_{iu}$. Finally, we can compute $P_{ac}$ which measures the probability that the model is accurate on its output given that it is confident in its prediction, and $P_{ui}$ which measures the probability that the model is uncertain about its output given that its prediction is wrong. They are respectively defined as 
\begin{equation}
    P_{ac}: p(\text{accurate }|\text{ certain}) = \frac{n_{ac}}{n_{ac} + n_{ic}}~,
\end{equation}
\begin{equation}
    P_{ui}: p(\text{uncertain }|\text{ inaccurate}) = \frac{n_{iu}}{n_{ic} + n_{iu}}~.
\end{equation}
Finally, $PAvPU$ computes the probability of the model being confident on accurate predictions and uncertain on inaccurate ones:
\begin{equation}
    PAvPU = \frac{n_{ac} + n_{iu}}{n_{ac} + n_{au} + n_{ic} + n_{iu}}~.
\end{equation}
These metrics can be calculated using various uncertainty thresholds, which define the meaning of \emph{certain} for the model. In this regard, we align with \cite{ddu_uncertainty, pavpu}: we use $w=4$ as the window size and 50\% as the threshold for defining a patch as accurate (given $w=4$, at least 9 out of 16 pixels in each patch must be correctly predicted by the model). Moreover, we estimate the uncertainty threshold as the average uncertainty of all pixels over the validation set.

The fourth metric is the \emph{Prediction Rejection Ratio} ($PRR$) \cite{PRR_deJorge}. It is obtained by rejecting samples with low confidence and by computing the accuracy vs the amount of rejected samples (\ie we compute the \emph{Rejection-Accuracy} curves), normalizing the area under the curve by that of an oracle and subtracting a baseline score with randomly sorted samples.
A model yielding high values for all four metrics can effectively differentiate between confident, accurate predictions and uncertain, inaccurate ones.

\cref{tab:uncertainty} presents uncertainty results for fault-free inference (first four columns) and under simulated application-level faults (last four columns). Our method consistently outperforms others in uncertainty estimation across all metrics, in both scenarios. Notably, ReLU6 clipping slightly reduces two metrics in fault-free conditions on Cityscapes.

%% file: tables_simulated_extended.tex
\begin{table*}[t]
    \begin{center}
    \small
    \setcellgapes{1.5pt}
    \makegapedcells
    \resizebox{\textwidth}{!}{
    \begin{NiceTabular}{| l | c | c | c c c | c || c | c | c c c | c |}
        \midrule
        
        {} & \multicolumn{6}{c}{\textbf{GTA5} \cite{richter2016gta5}} & \multicolumn{6}{c}{\textbf{Cityscapes} \cite{cordts2016cityscapes}} \\
        \cline{2-13}
        
        {} & \textbf{Fault-Free} & \multicolumn{5}{c}{\textbf{Fault-Injected}} & \textbf{Fault-Free} & \multicolumn{5}{c}{\textbf{Fault-Injected}} \\
        \cline{2-13}
        \vspace{12pt}
        {} & \multirow{2}{*}{mIoU (\%)}  & \multirow{2}{*}{mIoU (\%)}  & Masked   & No Impact & Tolerable  & Critical & \multirow{2}{*}{mIoU (\%)}  & \multirow{2}{*}{mIoU (\%)}  & Masked   & No Impact & Tolerable  & Critical  \\
        {} &   &    &  SDCs (\%) & SDCs (\%) & SDCs (\%) & SDCs (\%) &   &    &  SDCs (\%) & SDCs (\%) & SDCs (\%) & SDCs (\%) \\  
        \midrule
        
        No Hardening & 78.03 \scriptsize{$\pm$ 0.22} & 64.29 \scriptsize{$\pm$ 0.27} & 1.12 \scriptsize{$\pm$ 0.17} & 0.73 \scriptsize{$\pm$ 0.15} & 78.95 \scriptsize{$\pm$ 2.15} & 19.20 \scriptsize{$\pm$ 1.85} & \textbf{72.62} \scriptsize{$\pm$ 0.53} & 51.59 \scriptsize{$\pm$ 0.67} & 1.27 \scriptsize{$\pm$ 0.22} & 0.60 \scriptsize{$\pm$ 0.16} & 75.07 \scriptsize{$\pm$ 1.86} & 23.06 \scriptsize{$\pm$ 1.78} \\
        
        \midrule
        
        Fault-Aware Training\cite{cavagneroIOLTS2022} & \underline{78.05} \scriptsize{$\pm$ 0.28} & 70.78 \scriptsize{$\pm$ 0.43} & 1.14 \scriptsize{$\pm$ 0.17} & 0.71 \scriptsize{$\pm$ 0.15} & 83.72 \scriptsize{$\pm$ 1.92} & 14.43 \scriptsize{$\pm$ 1.63} & 72.30 \scriptsize{$\pm$ 0.79} & 56.14 \scriptsize{$\pm$ 0.61} & 1.00 \scriptsize{$\pm$ 0.19} & 0.37 \scriptsize{$\pm$ 0.14} & 75.66 \scriptsize{$\pm$ 1.84} & 22.97 \scriptsize{$\pm$ 1.79} \\
        ReLU6 \cite{cavagneroIOLTS2022, burel2024relu6} & 77.66 \scriptsize{$\pm$ 0.37} & 72.60 \scriptsize{$\pm$ 0.21} & 0.98 \scriptsize{$\pm$ 0.16} & 0.65 \scriptsize{$\pm$ 0.14} & 86.21 \scriptsize{$\pm$ 1.83} & 12.16 \scriptsize{$\pm$ 1.42} & 71.73 \scriptsize{$\pm$ 1.01} & 55.12 \scriptsize{$\pm$ 0.98} & 0.90 \scriptsize{$\pm$ 0.18} & 0.80 \scriptsize{$\pm$ 0.17} & 75.20 \scriptsize{$\pm$ 1.85} & 23.10 \scriptsize{$\pm$ 1.77} \\
        ReLU6 + Fault-Aware Training \cite{cavagneroIOLTS2022} & 77.83 \scriptsize{$\pm$ 0.30} & 73.25 \scriptsize{$\pm$ 0.47} & 1.03 \scriptsize{$\pm$ 0.16} & 0.65 \scriptsize{$\pm$ 0.14} & 87.42 \scriptsize{$\pm$ 1.74} & 10.90 \scriptsize{$\pm$ 1.31} & 72.14 \scriptsize{$\pm$ 0.60} & 58.36 \scriptsize{$\pm$ 0.73} & 0.93 \scriptsize{$\pm$ 0.18} & 0.80 \scriptsize{$\pm$ 0.17} & 76.42 \scriptsize{$\pm$ 1.82} & 21.85 \scriptsize{$\pm$ 1.81} \\
        
        \midrule

        AMMS \cite{amms_burel} & 78.02 \scriptsize{$\pm$ 0.25} & \underline{76.14} \scriptsize{$\pm$ 0.29} & 0.95 \scriptsize{$\pm$ 0.16} & 0.61 \scriptsize{$\pm$ 0.14} & 90.83 \scriptsize{$\pm$ 1.56} & \underline{7.61} \scriptsize{$\pm$ 1.09} & 72.47 \scriptsize{$\pm$ 0.53} & \underline{67.97} \scriptsize{$\pm$ 0.73} & 0.97 \scriptsize{$\pm$ 0.19} & 0.80 \scriptsize{$\pm$ 0.17} & 80.90 \scriptsize{$\pm$ 1.73} & \underline{17.33} \scriptsize{$\pm$ 1.89} \\
        
        \midrule
        
        ReLUMax (Ours) & \textbf{78.06} \scriptsize{$\pm$ 0.26} & \textbf{77.48} \scriptsize{$\pm$ 0.32} & 0.89 \scriptsize{$\pm$ 0.15} & 0.58 \scriptsize{$\pm$ 0.13} & 93.27 \scriptsize{$\pm$ 1.28} & \textbf{5.26} \scriptsize{$\pm$ 0.83} & \underline{72.53} \scriptsize{$\pm$ 0.72} & \textbf{70.07} \scriptsize{$\pm$ 0.83} & 1.00 \scriptsize{$\pm$ 0.19} & 0.60 \scriptsize{$\pm$ 0.16} & 85.40 \scriptsize{$\pm$ 1.65} & \textbf{13.00} \scriptsize{$\pm$ 1.95} \\
        
        \midrule
    
    \end{NiceTabular}}
    \end{center}
    \vspace{-5mm}
    \caption{Average fault-free and fault-injected mean Intersection over Union (mIoU), using DeepLabV3 on ResNet-50. Each experiment is repeated three times. We report also the standard deviation. For the Silent Data Corrupts (SDCs) results, we aggregate the number of observed SDCs over the three runs. For Critical SDCs, the lower the better. \textbf{Bold} indicates the best results. \underline{Underline} the second best.}
    \vspace{-3mm}
    \label{tab:sim}
\end{table*}

%% file: images_figures_maps_zoom.tex
\begin{figure*}[t]
\centering
    \includegraphics[width=0.75\linewidth]{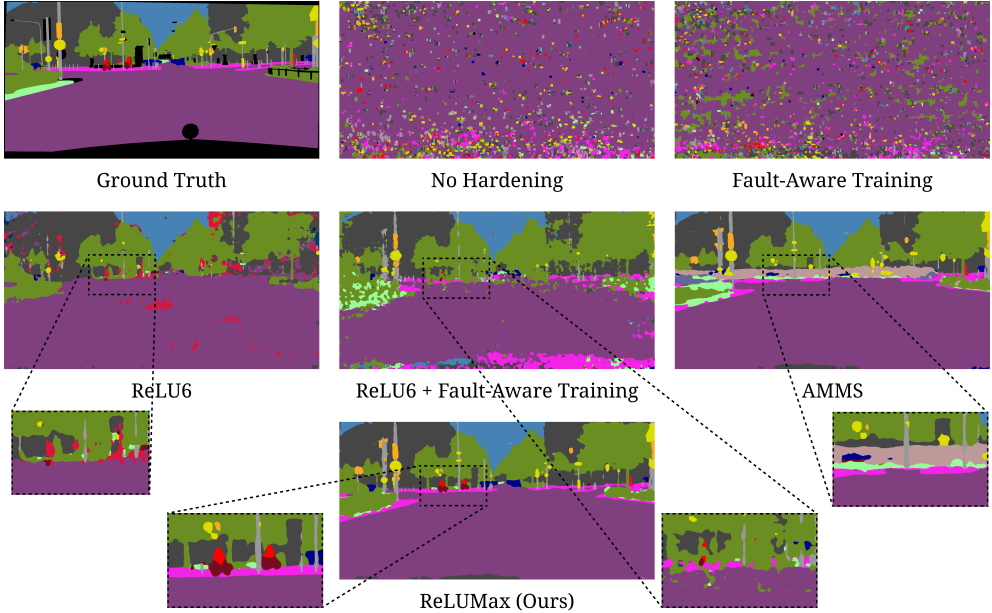}
    \vspace{-3mm}
    \caption{Segmentation maps predicted by the methods under assessment, when simulating injections on the validation split of the Cityscapes dataset. Each color represents a pixel-level annotation of the corresponding semantic class. The choice of the example to visualize is based on the worst recorded mean Intersection over Union (mIoU) when no hardening is introduced. For fairness of comparison, all the methods are injected deterministically in the same way, according to the simulated fault model provided by \cite{cavagneroIOLTS2022}.}\label{fig:maps_zoom}
\end{figure*}

%% file: tables_uncertainty.tex
\begin{table*}[ht]
    \begin{center}
    \small
    \setcellgapes{1.5pt}
    \makegapedcells
    \resizebox{\textwidth}{!}{
    \begin{NiceTabular}{| l | c c c | c | c c c | c || c c c | c | c c c | c | }
        \midrule
        
        {} & \multicolumn{8}{c}{\textbf{GTA5} \cite{richter2016gta5}} & \multicolumn{8}{c}{\textbf{Cityscapes} \cite{cordts2016cityscapes}} \\
        \cline{2-17}
        
        {} & \multicolumn{4}{c}{\textbf{Fault-Free}} & \multicolumn{4}{c}{\textbf{Fault-Injected}} & \multicolumn{4}{c}{\textbf{Fault-Free}} & \multicolumn{4}{c}{\textbf{Fault-Injected}}\\
        \cline{2-17}
        
        {} & $P_{ac}$ & $P_{ui}$ & $PAvPU$ & $PRR$ (\%) & $P_{ac}$ & $P_{ui}$ & $PAvPU$ & $PRR$ (\%) & $P_{ac}$ & $P_{ui}$ & $PAvPU$ & $PRR$ (\%) & $P_{ac}$ & $P_{ui}$ & $PAvPU$ & $PRR$ (\%) \\        
        \midrule
        
        No Hardening & 0.8896 & 0.6325 & 0.9977 & 1.58 & 0.8974 & 0.6389 & 0.9982 & 1.55 & 0.8852 & 0.6102 & 0.9922 & 1.27 & 0.8108 & 0.5719 & 0.9496 & 1.20 \\
        
        \midrule
        
        Fault-Aware Training \cite{cavagneroIOLTS2022} & 0.8902 & 0.6378 & 0.9985 & 1.63 & 0.9012 & 0.6442 & 0.9988 & 1.61 & 0.8830 & \underline{0.6148} & \underline{0.9949} & \underline{1.51} & 0.8216 & 0.6080 & 0.9599 & \underline{1.45} \\
        ReLU6 \cite{cavagneroIOLTS2022,burel2024relu6} & 0.8897 & 0.6356 & 0.9981 & 1.60 & 0.8998 & 0.6421 & 0.9986 & 1.59 & 0.8852 & 0.6128 & 0.9936 & 1.47 & 0.8485 & 0.6094 & 0.9867 & 1.43 \\
        ReLU6 + Fault-Aware Training \cite{cavagneroIOLTS2022} & 0.8899 & 0.6369 & 0.9983 & 1.62 & 0.9005 & 0.6435 & 0.9987 & 1.60 & 0.8840 & 0.6090 & 0.9938 & 1.39 & 0.8530 & 0.5964 & 0.9852 & 1.36 \\

        \midrule

        AMMS \cite{amms_burel} & \underline{0.8904} & \underline{0.6385} & \underline{0.9986} & \underline{1.64} & \underline{0.9018} & \underline{0.6449} & \underline{0.9989} & \underline{1.63} & \underline{0.8853} & 0.6105 & 0.9917 & 1.44 & \underline{0.8813} & \underline{0.6141} & \underline{0.9955} & \underline{1.45} \\
        
        \midrule
        

        ReLUMax (Ours) & \textbf{0.8906} & \textbf{0.6392} & \textbf{0.9988} & \textbf{4.85} & \textbf{0.9021} & \textbf{0.6456} & \textbf{0.9990} & \textbf{4.80} & \textbf{0.8863} & \textbf{0.6228} & \textbf{0.9993} & \textbf{8.28} & \textbf{0.8844} & \textbf{0.6257} & \textbf{0.9996} & \textbf{8.26} \\

        \midrule
    
    \end{NiceTabular}}
    \end{center}
    \vspace{-4mm}
    \caption{Uncertainty estimates of the hardening techniques applied to DeepLabV3 on ResNet-50. 
    The results are obtained via predictive uncertainty, computed using softmax entropy. We use a window size of 4x4, the accuracy threshold is set to 50\% and the uncertainty threshold is set as the average uncertainty of all pixels over the validation set. \textbf{Bold} indicates the best results. \underline{Underline} the second best.}
    \vspace{-3mm}
    \label{tab:uncertainty}
\end{table*}

%% file: sections_conclusions.tex
\vspace{-1.5mm}
\section{Conclusions}
\vspace{-1.5mm}

We investigated the robustness of semantic segmentation models to transient faults, evaluating existing hardening techniques under realistic fault injection scenarios. Moreover, we propose ReLUMax, a novel activation function that enhances model resilience by identifying acceptable activation ranges during training and clipping high-intensity faulty activations to zero during deployment. 
Despite its simplicity the proposed solution provides top accuracy results and shows promising performance in terms of maintaining model confidence. Up to our knowledge no previous work assessed uncertainty of hardened models, while we believe it is a crucial aspect to consider, especially in critical application scenarios as autonomous driving.

Future work will extend our study to further architectures (\ie transformer-based) and will involve tests on hardware platforms under neutron beam irradiation by following \cite{cavagneroIOLTS2022}.

\smallskip
\noindent\textbf{Acknowledgements.} 
\small{L.I. acknowledges the grant received from the European Union Next-GenerationEU (Piano Nazionale di Ripresa E Resilienza (PNRR)) DM 351 on Trustworthy AI. T.T. acknowledges the EU project ELSA - European Lighthouse on Secure and Safe AI.
This study was carried out within the FAIR - Future Artificial Intelligence Research and received funding from the European Union Next-GenerationEU (PIANO NAZIONALE DI RIPRESA E RESILIENZA (PNRR) – MISSIONE 4 COMPONENTE 2, INVESTIMENTO 1.3 – D.D. 1555 11/10/2022, PE00000013). 
This manuscript reflects only the authors’ views and opinions, neither the European Union nor the European Commission can be considered responsible for them.
}